\documentclass{article}

\PassOptionsToPackage{numbers, compress}{natbib}
\usepackage[preprint]{neurips_2026}

\usepackage[utf8]{inputenc} 
\usepackage[T1]{fontenc}    
\usepackage{hyperref}       
\usepackage{url}            
\usepackage{booktabs}       
\usepackage{amsfonts}       
\usepackage{nicefrac}       
\usepackage[nopatch=showhyphens]{microtype} 
\usepackage[table]{xcolor}  
\usepackage[most]{tcolorbox}
\usepackage{graphicx}
\usepackage{amsmath}
\usepackage{amssymb}
\usepackage{longtable}
\usepackage{placeins}
\usepackage{algorithm}
\usepackage{algpseudocode}

\definecolor{PaperBlue}{HTML}{2563EB}
\definecolor{PaperGray}{HTML}{8A8F98}
\definecolor{PaperDark}{HTML}{111827}
\definecolor{PaperPurple}{HTML}{7C3AED}
\hypersetup{
  colorlinks=true,
  linkcolor=PaperBlue,
  citecolor=PaperBlue,
  urlcolor=PaperBlue
}

\newcommand{\method}{Boundary-Aware Curriculum RL}

\title{Curriculum Reinforcement Learning Can Incentivize Reasoning Capacity in LLMs Beyond the Base Model}

\author{%
  \textbf{Pengxiang Cai}$^{1}$ \qquad
  \textbf{Tianchen Fang}$^{2}$ \qquad
  \textbf{Xiaohan Li}$^{1}$ \\[6pt]
  \textbf{Qingyuan Zeng}$^{1}$ \qquad
  \textbf{Guocong Li}$^{3}$ \qquad
  \textbf{Jintai Chen}$^{1\dag}$ \\[8pt]
  \normalfont
  $^{1}$The Hong Kong University of Science and Technology (Guangzhou) \\[2pt]
  \normalfont
  $^{2}$Nanyang Technological University
  \quad
  $^{3}$Zhejiang University \\[6pt]
  \normalfont
  $^{\dag}$Correspondence:
  \texttt{jintaiCHEN@hkust-gz.edu.cn}
}

\begin{document}

\maketitle

\begin{abstract}
Reinforcement learning with verifiable rewards (RLVR) is widely viewed as a promising path toward continuously improving large language models. Recent works, however, suggest that mainstream RLVR often reallocates sampling probabilities among trajectories already present in the base model: it can improve sampling efficiency, reflected by higher pass@1 scores, but yields limited gains, and can even decrease pass@$k$ scores when $k$ is large, and therefore may fail to expand the base model's reasoning capacity boundary. In this paper, we present a boundary-aware Curriculum RL approach to move beyond the base model's reasoning capacity boundary. Our approach first uses pass@$k$ sampling to locate the current reasoning capacity boundary, then applies targeted teacher guidance to examples near or beyond that boundary, and finally uses RL to consolidate the newly introduced reasoning patterns. Across Qwen, Llama, and DeepSeek base models, boundary-aware Curriculum RL improves both pass@1 scores and pass@256 scores, with pass@1 reflecting one-attempt performance and pass@256 serving as an empirical proxy for the reasoning capacity boundary. In our experiments, average pass@256 improves by 9.8 percentage points over the base models and by 10.3 percentage points over Vanilla RLVR. These results suggest that boundary-aware Curriculum RL can provide a scalable route for LLMs to continuously improve beyond the base model's empirical reasoning capacity boundary.
\end{abstract}

\section{Introduction}

The notable success of reasoning models such as OpenAI o1, DeepSeek-R1, and Kimi k1.5 has focused attention on how reinforcement-learning post-training can produce LLMs with long-chain deep reasoning abilities~\citep{openai2024openaia,deepseek-ai2025deepseekr1b,team2025kimi,comanici2025gemini,seed2025seed15thinking,yang2024qwen25matha,yang2025qwen3a,muennighoff2025s1a}. A major reason for this shift is the rise of reinforcement learning with verifiable rewards (RLVR): in mathematics, programming, and other outcome-verifiable domains, RL can optimize against automatically checkable answers rather than expensive human preferences~\citep{cobbe2021trainingb,chen2021evaluatinga,hendrycks2021measuringb,gao2024omnimatha,glazer2024frontiermatha,sun2025challenging,lu2023mathvista}. Algorithms and training recipes such as GRPO, DAPO, and GSPO further show that verifiable reward training can turn base or instruction-tuned models into substantially stronger mathematical reasoners~\citep{shao2024deepseekmathb,yu2025dapo,zheng2025groupa,su2025trusta,huang2025rzeroa,yan2025learninga,zuo2025ttrla}. These successes have encouraged an ambitious view of RLVR as a path toward continuously improving language models.

Most current RLVR pipelines are built around rollout-based policy optimization, with GRPO-style algorithms serving as a representative recipe. For each prompt, the model samples a group of candidate solutions; verifiable rewards identify correct and incorrect rollouts; and the policy update increases the likelihood of rewarded behavior while suppressing unrewarded behavior. This mechanism can improve answer accuracy without human preference labels, but it also raises a more precise question: \textit{does RLVR teach the model to solve problems that were previously unreachable, or does it mainly make the model more likely to sample correct trajectories that already existed in the base-model distribution?} The Limit-of-RLVR study evaluates RLVR-trained models with large-$k$ pass@$k$ and finds that standard RLVR often improves pass@1 while failing to expand, and sometimes narrowing, the base model's large-$k$ reasoning capacity boundary~\citep{yue2025doesa}. Related analyses suggest that RL post-training can amplify behaviors already learned before RL~\citep{zhao2025echoa}, and that group policy optimization can collapse into distribution sharpening unless rare correct trajectories receive sufficient learning pressure~\citep{he2025rewardinga}. We therefore analyze full pass@\(k\) curves in this work, with particular emphasis on large-\(k\) performance: if an RL-trained model improves pass@1 while its large-$k$ pass@$k$ remains close to, or below, the base model, then RL has mainly reallocated sampling probabilities within the base model's existing reasoning support; if previously unsolved problems become solvable under the same large sampling budget, then the reasoning capacity boundary has moved. This perspective explains why standard RLVR may oscillate inside the base model's inherited reasoning capacity boundary: it learns from sampled rollouts, so problems with no correct rollout provide no successful reasoning trace to reinforce. Meanwhile, policy sharpening can reduce sampling diversity and make some rare base-model solutions disappear at large $k$.

At the same time, we identify a group-level mechanism behind this limitation: extremely difficult examples can form zero-advantage groups after rollout. With binary outcome rewards, if every sampled solution receives reward 0, the group mean is also 0, so group-normalized advantages collapse to 0 for all rollouts. The example then contributes no reward-driven positive or negative preference signal to policy optimization.

This observation suggests that extremely difficult examples may receive little benefit from standard RLVR: if all sampled rollouts fail, the example provides no advantage signal for policy optimization. Yet these examples are precisely where the model's reasoning capacity boundary lies. Standard RLVR therefore lacks an effective mechanism for pushing beyond this reasoning capacity boundary, because the examples that most need new reasoning patterns are also the examples that provide no reward-driven learning signal. To make such examples usable, we first need to locate them. We use pass@$k$ sampling to distinguish problems that are already solvable, problems that are only barely reachable after many attempts, and problems that remain unsolved even after many attempts. This boundary-aware partition allows us to provide targeted guidance on examples for which standard RLVR would otherwise obtain no useful advantage signal.

After locating examples likely to form zero-advantage groups, the remaining challenge is to convert them into examples from which RL can learn. For problems outside the current reasoning capacity boundary, all-failed rollouts contain no successful trajectory, so RL alone has little basis for discovering the missing solution strategy. Some external structure is needed before reward learning can become informative. Full-scale teacher distillation provides such structure, and chain-of-thought distillation has long been used to transfer reasoning behavior from stronger teachers~\citep{hsieh2023distillinga,deepseek-ai2025deepseekr1b,chen2025unveilinga,li2025language,wen2025reasoninga,wu2025enhancinga}. However, distillation is teacher-dependent, often requires large corpora of teacher-generated reasoning traces, can be difficult to control, and may induce catastrophic forgetting when the data mixture is not carefully controlled. These tradeoffs motivate a localized-guidance design: the teacher supplies a small amount of structure near the reasoning capacity boundary, enough to turn examples that would otherwise form zero-advantage groups into training signal that RL can use, while RL remains responsible for consolidating the newly introduced reasoning patterns.

We implement this idea as boundary-aware Curriculum RL. Each curriculum round consists of three steps: (1) large-sample pass@$k$ sampling estimates the model's current reasoning capacity boundary; (2) targeted teacher guidance provides a small number of structured reasoning traces for examples near or beyond the reasoning capacity boundary; and (3) RL consolidation reinforces the newly introduced reasoning patterns into stable model behavior. The next round then begins with a fresh pass@$k$ exploration of the strengthened model, so the curriculum can target harder examples near the newly observed reasoning capacity boundary. In this way, the curriculum progressively turns examples near the reasoning capacity boundary that would otherwise form zero-advantage groups into trainable examples with useful reward signal, supporting reasoning capacity boundary expansion.

Our study leads to the following findings and contributions:
\begin{itemize}
  \item We identify a key blind spot of rollout-based RLVR: problems that remain unsolved under large-sample decoding are likely to form zero-advantage groups after rollout, producing no group-normalized advantage and thus little reward-driven learning signal. This observation motivates using large-$k$ pass@$k$ sampling to locate the model's current reasoning capacity boundary.
  \item We develop a reasoning capacity boundary exploration procedure based on large-$k$ pass@$k$ sampling. The procedure separates problems that are already solvable, barely reachable after many attempts, and still unreachable, and uses this reachability signal to construct model-specific curricula.
  \item We show how examples that typically form zero-advantage groups can be made trainable. By providing small, targeted teacher guidance near or beyond the reasoning capacity boundary, we supply sufficient structured signals to make previously failing examples reward-reachable, while RL consolidation stabilizes these newly acquired patterns into robust model behavior.
  \item We evaluate boundary-aware Curriculum RL on Qwen, Llama, and DeepSeek base models. Results show improvements from pass@1 to pass@256, indicating stronger one-attempt performance and expansion of the observed reasoning capacity boundary. This consistent pattern across model families supports the generality of our proposed boundary-aware Curriculum RL.
\end{itemize}

\begin{figure}[t]
  \centering
  \includegraphics[width=\linewidth]{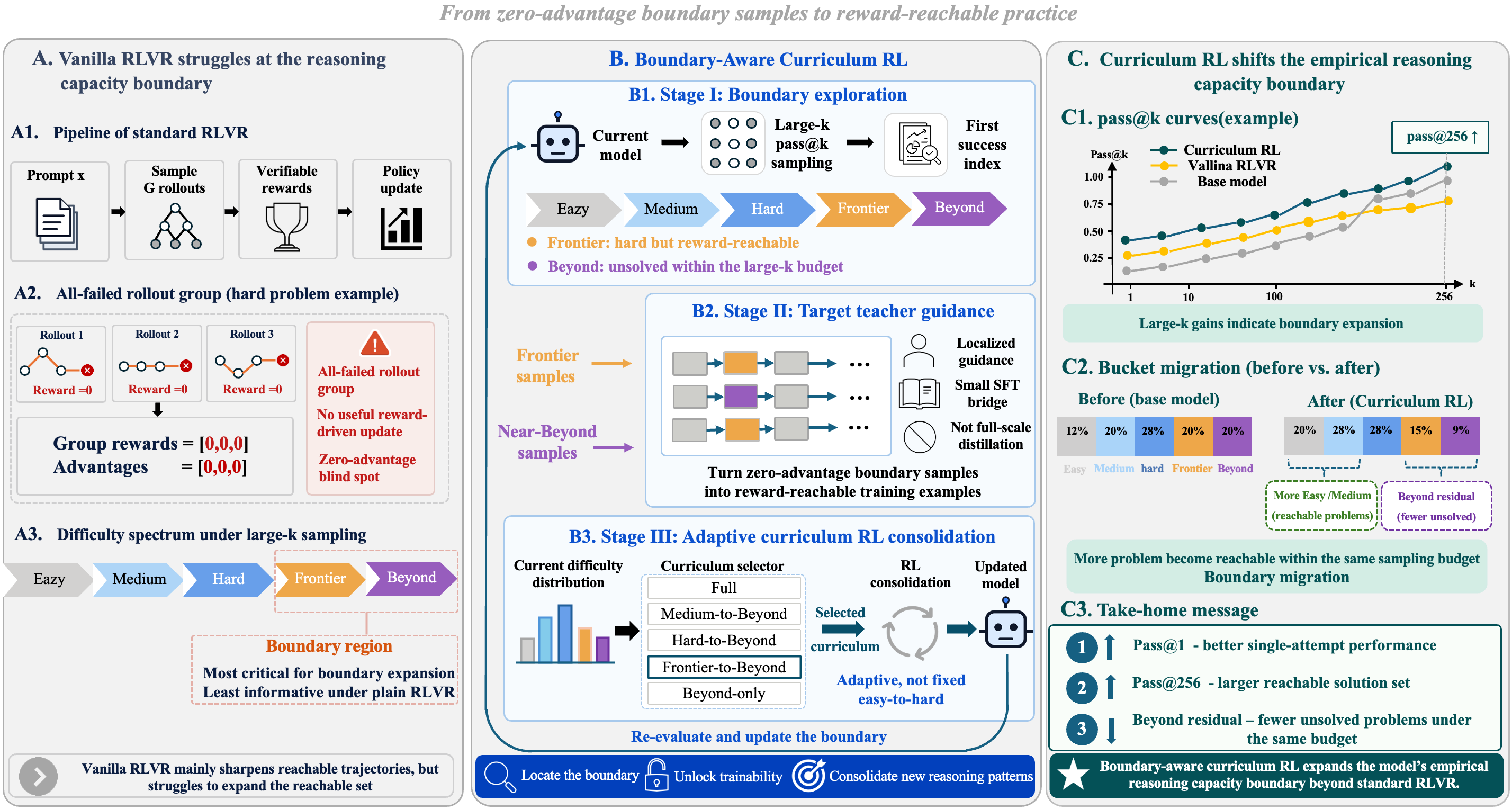}
  \caption{Overview of \method{}. \textbf{(A)} Vanilla RLVR can sharpen reachable trajectories but receives little reward-driven signal from all-failed rollout groups: with binary rewards, all rollouts obtain reward 0 and group-normalized advantages collapse to zero. \textbf{(B)} \method{} uses large-$k$ pass@$k$ sampling to locate the empirical reasoning capacity boundary, applies small targeted teacher guidance to examples near or beyond that boundary, and then uses RL consolidation to make the introduced reasoning patterns stable. \textbf{(C)} The resulting curriculum shifts the empirical reasoning capacity boundary, as reflected by improved pass@1/pass@256 behavior and fewer problems remaining unsolved under the same sampling budget.}
  \label{fig:overview}
\end{figure}

\section{Related Work}

\paragraph{Limits of RLVR.}
Recent analyses examine whether RLVR can improve a model's reasoning capacity beyond the base model, or whether it mainly changes the probability of sampling existing trajectories. Limit-of-RLVR uses large-$k$ pass rates, coverage, and perplexity to argue that current RLVR often remains bounded by the base model~\citep{yue2025doesa}. Related work suggests that RL post-training can amplify behaviors learned before RL~\citep{zhao2025echoa}, and that GRPO-like optimization can degenerate into distribution sharpening when rare correct trajectories receive insufficient learning pressure~\citep{he2025rewardinga}. Other studies give a more optimistic view: RLVR can incentivize correct reasoning when evaluated with trajectory-sensitive metrics~\citep{wen2025reinforcementa,zheng2024processbencha}, and longer, carefully controlled RL can expand reasoning capacity boundaries~\citep{liu2025prorl}.

\paragraph{Reasoning distillation and teacher guidance.}
Teacher distillation transfers reasoning behavior from stronger teachers to weaker students, often through chain-of-thought rationales or long reasoning traces~\citep{hsieh2023distillinga,deepseek-ai2025deepseekr1b,chen2025unveilinga,li2025language,wen2025reasoninga,wu2025enhancinga}. Such guidance can introduce reasoning patterns that sparse outcome rewards may not reveal, but full-scale distillation makes the training outcome depend on teacher quality, the scale of teacher-generated reasoning traces, and data mixture design. More broadly, continual fine-tuning studies show that adding new specialized behavior can cause forgetting of earlier knowledge and reasoning skills when the data mixture is not controlled~\citep{luo2025empiricala,huang2024mitigatinga,huang2025mitigatinga,jiang2024unlocking,wang2024inscla,wang2025continuala,zheng2025spuriousa}.

\paragraph{Curriculum learning for reasoning.}
Recent LLM reasoning work studies how training examples should be selected and ordered during RL. Reverse curriculum RL constructs easier subproblems by starting from later parts of a reasoning process~\citep{xi2024traininga}; easy-to-hard curriculum RL improves reasoning by staging task difficulty~\citep{parashar2025curriculum,zhao2024automatic,amani2025rl,lai2026taclera,zeng2025cures}; online difficulty filtering emphasizes that RL data should remain neither too easy nor too hard~\citep{bae2026online}; and SimpleRL-Zoo shows that zero-RL behavior depends strongly on model identity and query difficulty~\citep{zeng2025simplerlzooa}. These works highlight the importance of example selection, while our setting uses large-$k$ pass@$k$ sampling to define difficulty through each model's observed reasoning capacity boundary.

\section{Method}

\method{} turns examples near the empirical reasoning capacity boundary into trainable RL data. Each curriculum round has three steps: (1) large-sample pass@$k$ sampling estimates the current reasoning capacity boundary; (2) targeted teacher guidance adds a small number of structured traces for examples near or just beyond that boundary; and (3) based on the pass@$k$ exploration results, we choose a suitable-difficulty curriculum for RL, allowing the model to consolidate the newly learned reasoning patterns. The next round starts with a new pass@$k$ exploration of the updated model and then targets the newly observed reasoning capacity boundary.

\paragraph{Zero-advantage groups.}
The need for reasoning capacity boundary exploration follows from the Limit-of-RLVR observation that current RLVR often improves pass@1 by shifting sampling probabilities toward reasoning paths already available in the base model, while large-$k$ pass@$k$ coverage remains close to, or below, the base model~\citep{yue2025doesa}. Based on this observation, we identify a concrete failure mode: problems that the current policy cannot solve in many rollouts provide no successful trajectory from which outcome-reward RL can learn.

This limitation has a simple group-level form. For a prompt \(x\), suppose a GRPO-style update samples \(G\) rollouts \(\{y_i\}_{i=1}^G\) with binary rewards \(r_i\in\{0,1\}\), and uses the group-normalized advantage
\[
  \hat A_i = \frac{r_i-\bar r}{\sigma_r+\epsilon},
  \qquad
  \bar r = \frac{1}{G}\sum_{j=1}^G r_j .
\]
Here, \(\sigma_r\) is the standard deviation of rewards within the group. If all sampled rollouts fail, then \(r_i=0\) for every \(i\), so \(\bar r=0\), \(\sigma_r=0\), and \(\hat A_i=0\) for all rollouts. The reward-driven policy-gradient contribution for this prompt is therefore zero:
\[
  \sum_{i=1}^G \hat A_i \nabla_\theta \log \pi_\theta(y_i\mid x) = 0 .
\]
The example provides neither positive nor negative preference among sampled trajectories; for the reward-driven term, it becomes inert training data. Examples that form such zero-advantage groups are also candidates that define the current observed reasoning capacity boundary. Standard RLVR therefore faces an optimization dead zone: the examples most needed for reasoning capacity boundary expansion are precisely the examples least able to provide reward-driven learning signal after rollout. The first stage of \method{} uses large-sample pass@$k$ sampling to locate this region before applying targeted guidance.
Appendix~\ref{sec:appendix-zero-advantage} provides a more general form of this zero-advantage argument and relates it to reward-reachable groups.

\subsection{Stage I: Reasoning Capacity Boundary Exploration}

The first stage assigns each problem a model-specific difficulty label using a 256-sample decoding budget. For each model setting, we evaluate every problem with three decoding seeds, \(\{2027,2028,2029\}\), and record the first sample index at which a correct answer appears. If no correct answer appears within 256 samples, the problem is treated as unsolved under the current model. We average the first-success index across seeds to estimate how far the problem lies from the current model under the same sampler. The resulting pass@$k$ profile defines five difficulty levels:
\begin{tcolorbox}[
  colback=PaperBlue!3,
  colframe=PaperGray!70,
  boxrule=0.4pt,
  arc=1pt,
  left=3pt,
  right=3pt,
  top=2pt,
  bottom=2pt,
  boxsep=0pt,
  before skip=4pt,
  after skip=4pt,
  halign=center
]
\footnotesize
\begin{tabular}{@{}l@{\hspace{1.45em}}l@{\hspace{1.45em}}l@{\hspace{1.45em}}l@{\hspace{1.45em}}l@{}}
\textbf{Easy}:\,\(k=1\) &
\textbf{Medium}:\,\(2 \le k \le 8\) &
\textbf{Hard}:\,\(8 < k \le 32\) &
\textbf{Frontier}:\,\(32 < k \le 256\) &
\textbf{Beyond}:\,\(k > 256\)
\end{tabular}
\end{tcolorbox}
The Beyond group contains problems not solved within 256 samples and is therefore the region most likely to form zero-advantage groups under outcome-reward RL; the Frontier group contains hard but reward-reachable problems, matching the edge-of-competence condition emphasized in recent controlled studies~\citep{zhang2025interplaya}. This distinction prevents the curriculum from only repeating easy examples: easy examples mainly increase the sampling probability of already reachable solutions, whereas Frontier and near-boundary Beyond examples directly target the reasoning capacity boundary and expose a training region that standard RLVR would otherwise leave unused.

\subsection{Stage II: Targeted Teacher Guidance}

The second stage turns reasoning capacity boundary exploration into trainable signal. We first extract examples near the reasoning capacity boundary for customized teacher guidance. Concretely, we draw 10 Frontier problems whose first-success index is closest to the pass@32 cutoff and 10 Beyond problems closest to the pass@256 cutoff, while maintaining dataset coverage and avoiding repeated samples; the full selection procedure is given in Appendix~\ref{sec:appendix-sft-selection}. These examples are neither trivial nor arbitrarily impossible: Frontier supplies hard-but-reward-reachable practice, while near-boundary Beyond examples supply targets just outside the current reasoning capacity boundary.

We then construct structured teacher traces for the selected Frontier/Beyond examples and use them as a small SFT bridge before RL. The teacher prompt, shown in Appendix~\ref{sec:appendix-sft-prompt}, asks for logical backbones, explicit self-reflection, multi-path exploration, and coherence checks. The small customized teacher guidance in \method{} provides enough local structure near the reasoning capacity boundary to make previously unreachable reasoning patterns visible to the optimizer. After this bridge, examples that would have formed zero-advantage groups can become reward-reachable, allowing RL to reinforce and stabilize the newly introduced reasoning patterns. This design separates \method{} from both standard outcome-reward RLVR and large-scale teacher distillation.

\subsection{Stage III: Curriculum Selection and RL Consolidation}

The third stage chooses the next RL training range from the pass@$k$ difficulty distribution and then applies standard GRPO to consolidate the reasoning patterns introduced in Stage II. We use five curriculum ranges defined over the Stage-I difficulty groups:
\begin{tcolorbox}[
  colback=PaperBlue!3,
  colframe=PaperGray!70,
  boxrule=0.4pt,
  arc=1pt,
  left=5pt,
  right=5pt,
  top=3pt,
  bottom=3pt,
  boxsep=0pt,
  before skip=4pt,
  after skip=4pt,
  halign=center
]
\small
\begin{tabular}{@{}l@{\hspace{0.95em}}l@{\hspace{0.95em}}l@{\hspace{0.95em}}l@{\hspace{0.95em}}l@{}}
\textbf{Easy-to-Beyond} &
\textbf{Medium-to-Beyond} &
\textbf{Hard-to-Beyond} &
\textbf{Frontier-to-Beyond} &
\textbf{Beyond-only}
\end{tabular}
\end{tcolorbox}
Easy-to-Beyond uses all five difficulty groups from Stage I; Medium-to-Beyond removes Easy examples; Hard-to-Beyond trains on Hard, Frontier, and Beyond; Frontier-to-Beyond trains on Frontier and Beyond; and Beyond-only trains only on Beyond examples.

Our curriculum selection follows a simple principle: retain enough reward-reachable examples while moving training toward the reasoning capacity boundary. For a candidate curriculum range \(\mathcal{C}\), define
\[
  M(\mathcal{C})=\sum_{b\in \mathcal{C}\cap R}p_b,
  \qquad
  H(\mathcal{C})=\sum_{b\in \mathcal{C}\cap B}p_b,
\]
where \(p_b=N_b/N\) is the pass@$k$ difficulty share, \(R=\{\mathrm{Medium},\mathrm{Hard},\mathrm{Frontier}\}\) contains non-trivial reward-reachable groups, and \(B=\{\mathrm{Frontier},\mathrm{Beyond}\}\) contains reasoning-capacity-boundary groups. We choose a curriculum that keeps \(M(\mathcal{C})\) sufficiently large and, among such candidates, prefers larger \(H(\mathcal{C})\). If Medium or Hard examples dominate, many problems remain solvable after repeated attempts but are not yet reliably solved, so RL starts from Medium-to-Beyond or Hard-to-Beyond. If Beyond examples dominate, training only on Beyond would produce many all-failed rollout groups, so we use a wider range such as Easy-to-Beyond or Medium-to-Beyond to keep useful reward signal in RL batches. Once Easy/Medium examples have largely been consolidated and the remaining difficulty concentrates near Frontier/Beyond, the curriculum moves toward Frontier-to-Beyond or Beyond-only.

After the curriculum range is chosen, we run standard GRPO on that data range. The next curriculum round begins with a new pass@$k$ exploration of the updated model, so the training range can move with the newly observed reasoning capacity boundary. In this work, we run three rounds of Stage I--II--III curriculum learning. Appendix~\ref{sec:appendix-curriculum-selection} reports the curriculum choices used in our experiments. We use this simple boundary-aware Curriculum RL procedure to evaluate whether RLVR can move beyond the base model's reasoning capacity boundary; optimal curriculum selection is left outside the scope of this work.

\begin{figure}[!htbp]
  \centering
  \includegraphics[width=\linewidth]{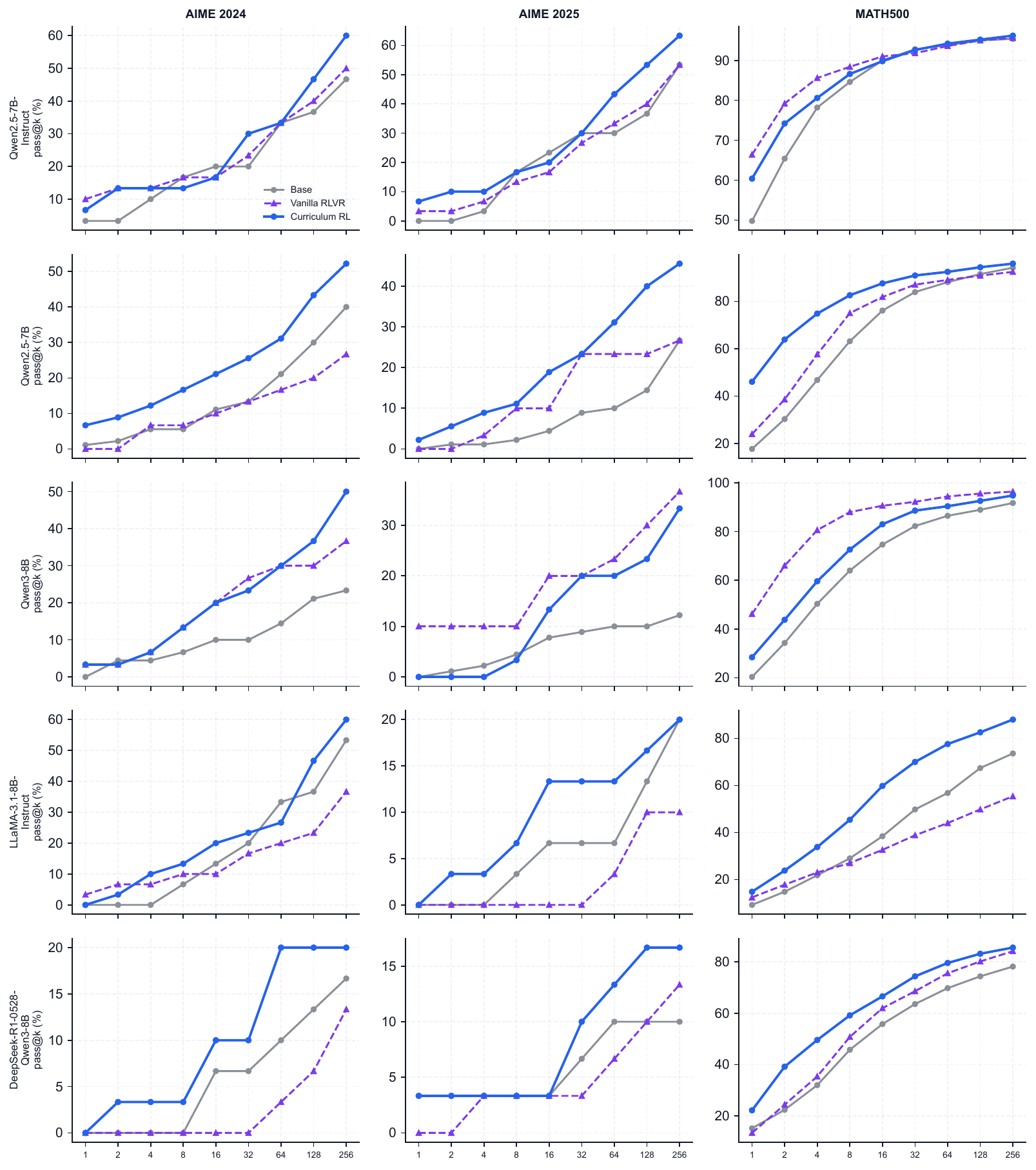}
  \caption{Full pass@$k$ curves across all model--benchmark pairs. Gray, purple dashed, and blue denote the base model, the Vanilla RLVR model, and the reported boundary-aware Curriculum RL model, respectively. Improvements near \(k=256\) indicate that more problems become solvable under a large sampling budget, rather than only being sampled more easily at small \(k\).}
  \label{fig:full-passk-grid}
\end{figure}

\section{Experiments}

\subsection{Experimental Setup}

We evaluate \method{} across five model settings: Qwen2.5-7B, Qwen2.5-7B-Instruct, Qwen3-8B, DeepSeek-R1-0528-Qwen3-8B, and Meta-Llama-3.1-8B-Instruct. For each setting, we compare three models: the base model, the Vanilla RLVR model, and the boundary-aware Curriculum RL model reported by our method. We train on a subset of DAPO-Math-17K and evaluate on AIME 2024, AIME 2025, and MATH500.

Following the large-sample evaluation protocol used in recent RLVR boundary analyses~\citep{yue2025doesa}, we report pass@$k$ for \(k\in\{1,2,4,8,16,32,64,128,256\}\). A pass@$k$ score measures whether the model can produce at least one correct answer within \(k\) sampled attempts. Thus, pass@1 measures one-attempt performance, while pass@256 measures whether a problem becomes solvable when the model is allowed a large sampling budget. We use pass@256 as an empirical proxy for the model's observed reasoning capacity boundary.

\subsection{Experimental Results}

We first inspect the full pass@$k$ curves because they show where along the sampling budget training changes model behavior. Figure~\ref{fig:full-passk-grid} compares the base model, the Vanilla RLVR model, and the boundary-aware Curriculum RL model on every model--benchmark pair. If training only made already available solutions easier to sample, most gains would appear near pass@1, with little change near pass@256. This is often what we observe for Vanilla RLVR: its large-\(k\) gains are limited, and in some cases pass@256 decreases relative to the base model. In contrast, boundary-aware Curriculum RL often improves both the low-\(k\) and high-\(k\) parts of the curve. The right side of the curve is the key evidence for reasoning capacity boundary expansion: improvements near \(k=256\) mean that more problems become solvable under the same large sampling budget.

\begin{figure}[t]
  \centering
  \includegraphics[width=\linewidth]{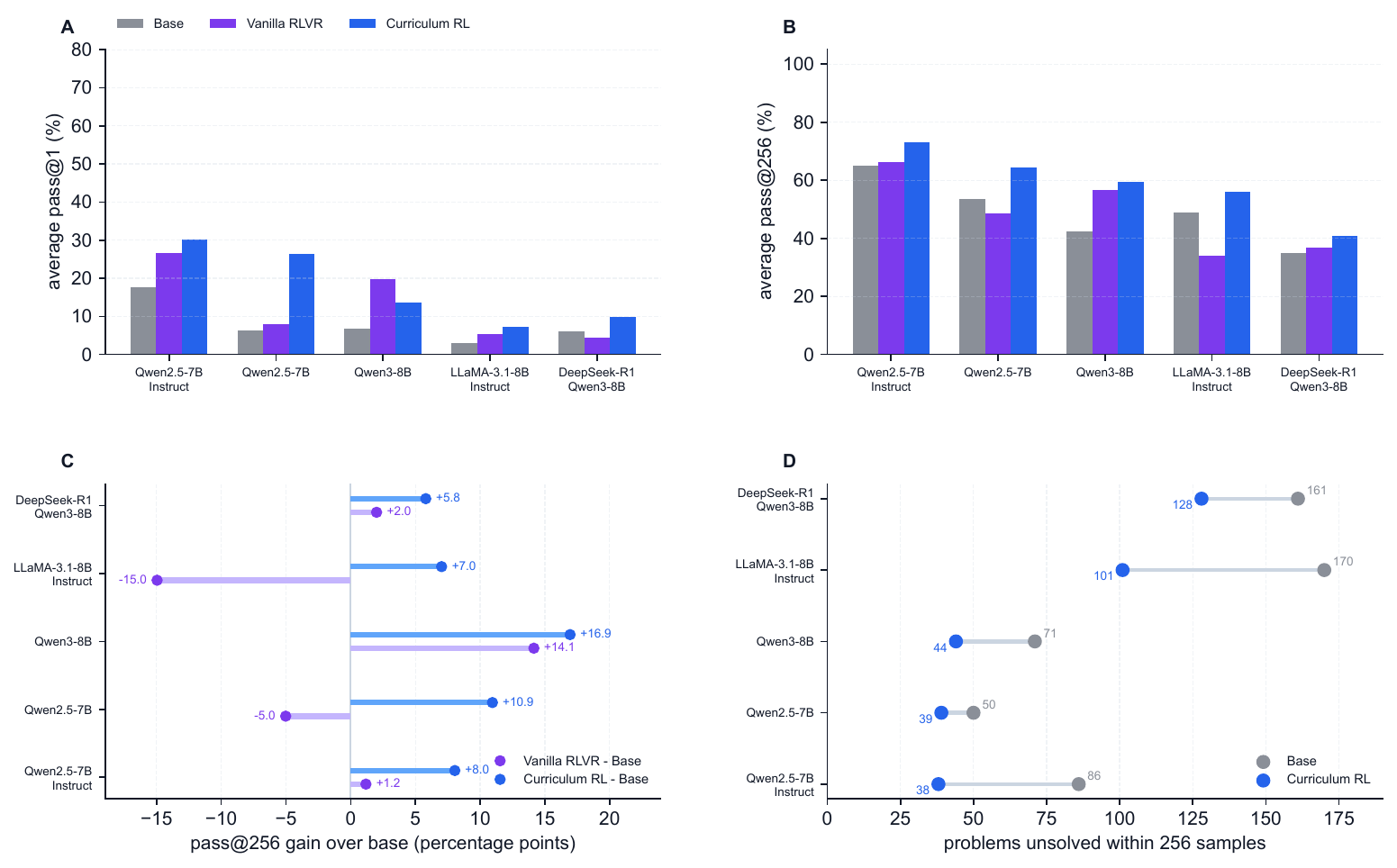}
  \caption{Summary of the main evaluation results. \textbf{A}, average pass@1 scores over AIME 2024, AIME 2025, and MATH500. \textbf{B}, average pass@256 scores over the same benchmarks. \textbf{C}, pass@256 gains over the base model for the Vanilla RLVR model and the boundary-aware Curriculum RL model. \textbf{D}, number of evaluation problems that remain unsolved within 256 sampled attempts for the base model and the boundary-aware Curriculum RL model.}
  \label{fig:main-results-summary}
\end{figure}

Figure~\ref{fig:main-results-summary} summarizes the averaged trends across AIME 2024, AIME 2025, and MATH500. Panel A shows that Curriculum RL improves pass@1 in every model setting, indicating better one-attempt performance. Panel B shows that the improvement also appears at pass@256, so the gain is not confined to low sampling budgets. Panel C uses the base model as a shared reference point: Vanilla RLVR changes mean pass@256 by \(-0.5\) percentage points on average, whereas Curriculum RL improves mean pass@256 by \(+9.8\) percentage points. Because pass@256 measures whether a problem becomes solvable within 256 sampled attempts, these results indicate that boundary-aware Curriculum RL makes more problems reachable under a large sampling budget.

\begin{table}[!htbp]
\caption{Benchmark-level pass@1 and pass@256 values for AIME 2024, AIME 2025, and MATH500. Figure~\ref{fig:main-results-summary} averages these values across benchmarks; this table keeps the per-benchmark results visible. Our method is highlighted in blue only when it is best or tied for best for a given benchmark and metric.}
\label{tab:main-pass}
\centering
\tiny
\setlength{\tabcolsep}{2.2pt}
\renewcommand{\arraystretch}{1.12}
\begin{tabular}{@{}p{0.24\linewidth}p{0.11\linewidth}cccccc@{}}
\toprule
Model & Benchmark &
\multicolumn{3}{c}{pass@1} &
\multicolumn{3}{c}{pass@256} \\
\cmidrule(lr){3-5}\cmidrule(l){6-8}
& & Base & Vanilla RLVR & Curriculum RL & Base & Vanilla RLVR & Curriculum RL \\
\midrule
Qwen2.5-7B-Instruct & AIME 2024 & 3.3 & 10.0 & \textcolor{PaperBlue}{\textbf{10.0}} & 46.7 & 50.0 & \textcolor{PaperBlue}{\textbf{60.0}} \\
 & AIME 2025 & 0.0 & 3.3 & \textcolor{PaperBlue}{\textbf{13.3}} & 53.3 & 53.3 & \textcolor{PaperBlue}{\textbf{63.3}} \\
 & MATH500 & 49.8 & 66.4 & \textcolor{PaperBlue}{\textbf{67.0}} & 95.4 & 95.6 & \textcolor{PaperBlue}{\textbf{96.2}} \\
\midrule
Qwen2.5-7B & AIME 2024 & 1.1 & 0.0 & \textcolor{PaperBlue}{\textbf{10.0}} & 40.0 & 26.7 & \textcolor{PaperBlue}{\textbf{52.2}} \\
 & AIME 2025 & 0.0 & 0.0 & \textcolor{PaperBlue}{\textbf{5.6}} & 26.7 & 26.7 & \textcolor{PaperBlue}{\textbf{45.6}} \\
 & MATH500 & 17.7 & 24.0 & \textcolor{PaperBlue}{\textbf{63.9}} & 94.1 & 92.4 & \textcolor{PaperBlue}{\textbf{95.9}} \\
\midrule
Qwen3-8B & AIME 2024 & 0.0 & 3.3 & \textcolor{PaperBlue}{\textbf{3.3}} & 23.3 & 36.7 & \textcolor{PaperBlue}{\textbf{50.0}} \\
 & AIME 2025 & 0.0 & 10.0 & 3.3 & 12.2 & 36.7 & 33.3 \\
 & MATH500 & 20.3 & 46.2 & 34.4 & 91.7 & 96.4 & 94.8 \\
\midrule
LLaMA-3.1-8B-Instruct & AIME 2024 & 0.0 & 3.3 & \textcolor{PaperBlue}{\textbf{6.7}} & 53.3 & 36.7 & \textcolor{PaperBlue}{\textbf{60.0}} \\
 & AIME 2025 & 0.0 & 0.0 & \textcolor{PaperBlue}{\textbf{0.0}} & 20.0 & 10.0 & \textcolor{PaperBlue}{\textbf{20.0}} \\
 & MATH500 & 9.2 & 12.4 & \textcolor{PaperBlue}{\textbf{15.4}} & 73.6 & 55.4 & \textcolor{PaperBlue}{\textbf{88.0}} \\
\midrule
DeepSeek-R1-0528-Qwen3-8B & AIME 2024 & 0.0 & 0.0 & \textcolor{PaperBlue}{\textbf{0.0}} & 16.7 & 13.3 & \textcolor{PaperBlue}{\textbf{20.0}} \\
 & AIME 2025 & 3.3 & 0.0 & \textcolor{PaperBlue}{\textbf{3.3}} & 10.0 & 13.3 & \textcolor{PaperBlue}{\textbf{16.7}} \\
 & MATH500 & 15.2 & 13.4 & \textcolor{PaperBlue}{\textbf{26.2}} & 78.2 & 84.2 & \textcolor{PaperBlue}{\textbf{85.6}} \\
\bottomrule
\end{tabular}
\end{table}

Table~\ref{tab:main-pass} reports the benchmark-level values behind these averaged trends. The table shows that the averaged gains are not driven by a single benchmark: Curriculum RL improves pass@256 over the base model on nearly all model--benchmark pairs, with especially clear gains on AIME 2024 and AIME 2025. The table also shows the instability of Vanilla RLVR: across model--benchmark pairs, it sometimes improves pass@256 but sometimes reduces pass@256 below the base model; on average, its pass@256 is slightly lower than that of the base model. This per-benchmark view supports the same conclusion as Figure~\ref{fig:main-results-summary}: boundary-aware Curriculum RL improves large-\(k\) behavior more consistently than Vanilla RLVR.

We next ask whether the large-\(k\) improvement corresponds to changes on individual problems. Table~\ref{tab:boundary-migration} counts evaluation problems that remain unsolved within 256 sampled attempts. Across all five model settings, this count decreases after Curriculum RL, falling from 538 to 350 in total. Among the 538 problems that were unsolved by the base models under the 256-sample budget, 226 become solvable after curriculum training. This problem-level analysis is stricter than pass@1 because it asks whether a problem that was outside the observed reasoning capacity boundary can be moved back inside that boundary.

\begin{table}[!htbp]
\caption{Problem-level reasoning capacity boundary migration. Counts are measured under the 256-sample budget. ``Base-unsolved solved'' counts original base-model unsolved problems that become solvable after Curriculum RL.}
\label{tab:boundary-migration}
\centering
\scriptsize
\setlength{\tabcolsep}{6pt}
\renewcommand{\arraystretch}{1.12}
\begin{tabular}{@{}p{0.38\linewidth}ccc@{}}
\toprule
Model & Base unsolved & Curriculum unsolved & Base-unsolved solved \\
\midrule
Qwen2.5-7B-Instruct & 86 & \textcolor{PaperPurple}{\textbf{38}} & \textcolor{PaperPurple}{\textbf{53/86}} \\
Qwen2.5-7B & 50 & \textcolor{PaperPurple}{\textbf{39}} & \textcolor{PaperPurple}{\textbf{15/50}} \\
Qwen3-8B & 71 & \textcolor{PaperPurple}{\textbf{44}} & \textcolor{PaperPurple}{\textbf{29/71}} \\
LLaMA-3.1-8B-Instruct & 170 & \textcolor{PaperPurple}{\textbf{101}} & \textcolor{PaperPurple}{\textbf{90/170}} \\
DeepSeek-R1-0528-Qwen3-8B & 161 & \textcolor{PaperPurple}{\textbf{128}} & \textcolor{PaperPurple}{\textbf{39/161}} \\
\bottomrule
\end{tabular}
\end{table}

\begin{figure}[t]
  \centering
  \includegraphics[width=\linewidth]{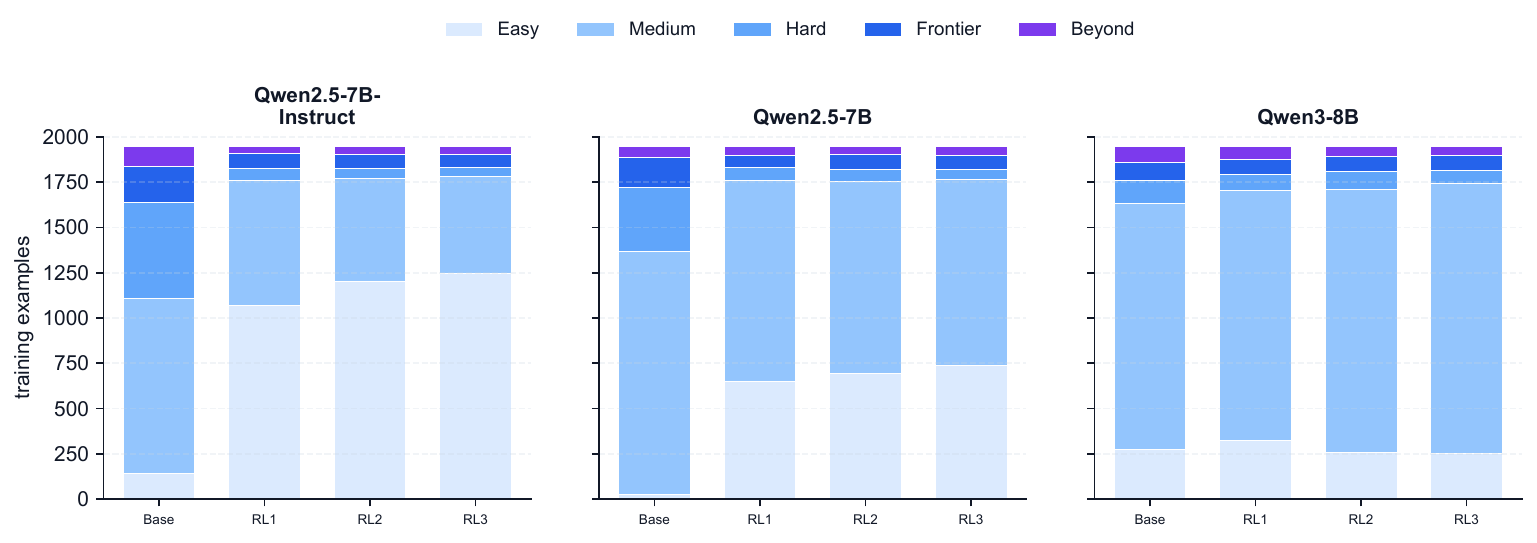}
  \caption{Expansion of training-example regions that provide useful reward-driven signal across curriculum rounds. We record the difficulty-group assignments produced by each pass@$k$ exploration step in the curriculum and summarize how training examples are distributed across Easy, Medium, Hard, Frontier, and Beyond from the base model to RL3. Fewer Beyond examples and more reward-reachable examples indicate that the curriculum expands the region from which RL can obtain useful reward-driven signal.}
  \label{fig:eps-stacks}
\end{figure}

Finally, Figure~\ref{fig:eps-stacks} examines how the curriculum changes the composition of training examples that can provide useful reward-driven signal for RL. We record the difficulty-group assignments produced by each pass@$k$ exploration step in the curriculum and summarize how training examples are distributed across Easy, Medium, Hard, Frontier, and Beyond. This view shows which examples can provide useful reward-driven signal for the next RL stage. Non-Beyond examples are reward-reachable under the 256-sample budget, whereas Beyond examples are more likely to form all-failed rollout groups. Across the three curriculum rounds, the Beyond region shrinks and more examples move into reward-reachable groups, indicating that the curriculum expands the set of examples from which RL can learn. This provides a mechanism-level diagnostic that boundary-aware Curriculum RL converts part of the zero-advantage region into reward-reachable examples. The exact curriculum choices used in our experiments are reported in Appendix~\ref{sec:appendix-curriculum-selection}, and additional training-example diagnostics are collected in Appendix~\ref{sec:appendix-figures}.

\section{Discussion and Conclusion}

\paragraph{From zero-advantage groups to reward-reachable examples.}
The Limit-of-RLVR observation motivates our setting: standard RLVR often improves pass@1 by reallocating sampling probabilities among trajectories already reachable from the base model, while the large-\(k\) reasoning capacity boundary often does not expand~\citep{yue2025doesa}. We focus on the group-level reason behind this failure mode. If all rollouts for a difficult example fail, group-normalized advantages become zero, leaving no useful reward-driven signal. \method{} addresses this by using pass@$k$ exploration to locate examples near the boundary, applying small targeted teacher guidance to make some of them reward-reachable, and then using RL to stabilize the new reasoning patterns. Figure~\ref{fig:eps-stacks} supports this mechanism: across curriculum rounds, fewer examples remain Beyond and more examples move into groups that can provide useful reward-driven signal.

\paragraph{Scope and limitations.}
The main finding of this work is that a boundary-aware training process can expand the empirical reasoning capacity boundary by converting zero-advantage groups into reward-reachable examples. We view \method{} as one instantiation of a broader boundary-aware training principle rather than a fixed recipe. This principle can be extended with different boundary estimators, guidance sources, curriculum rules, or RL optimizers, as long as the training process locates examples near the reasoning capacity boundary and makes otherwise zero-advantage groups trainable. We do not explore optimal curriculum selection in this work; instead, we focus on showing that this boundary-aware strategy can move the empirical reasoning capacity boundary.


\bibliographystyle{plainnat}
\bibliography{references}

\newpage
\appendix

\section{Supplementary Experimental Figures}
\label{sec:appendix-figures}
\begin{figure}[p]
  \centering
  \includegraphics[width=\linewidth]{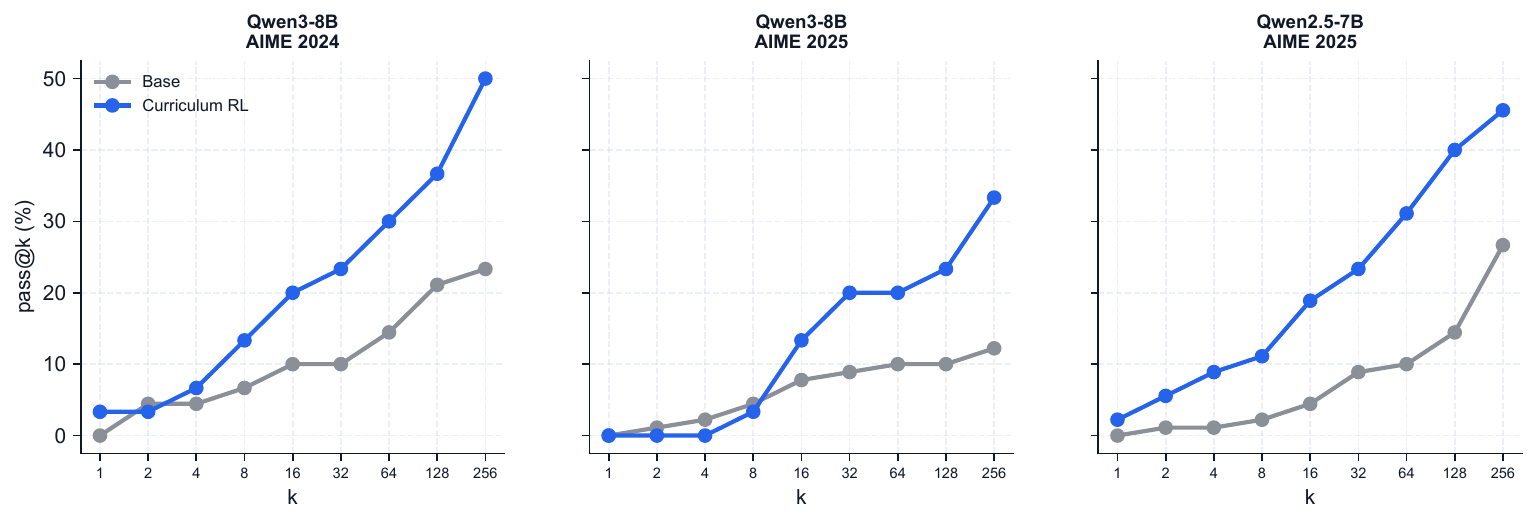}
  \caption{Representative pass@$k$ curves for three challenging model--benchmark pairs. These panels give a compact view of the large-\(k\) behavior shown more comprehensively in Figure~\ref{fig:full-passk-grid}.}
  \label{fig:passk-curves}
\end{figure}

\begin{figure}[p]
  \centering
  \includegraphics[width=\linewidth]{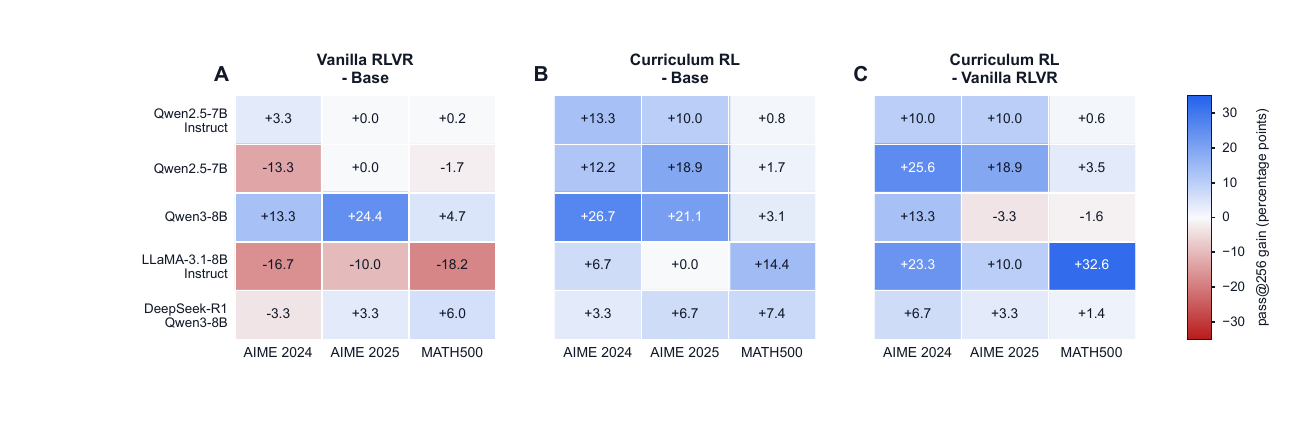}
  \caption{Per-benchmark pass@256 gain maps. Each cell reports the percentage-point change in pass@256 for one model--benchmark pair. Vanilla RLVR has mixed large-\(k\) effects, including several decreases relative to the base model. Curriculum RL shows broader positive pass@256 gains, indicating that more problems become solvable within the same 256-sample budget.}
  \label{fig:pass256-gain-heatmap}
\end{figure}

\begin{figure}[p]
  \centering
  \includegraphics[width=\linewidth]{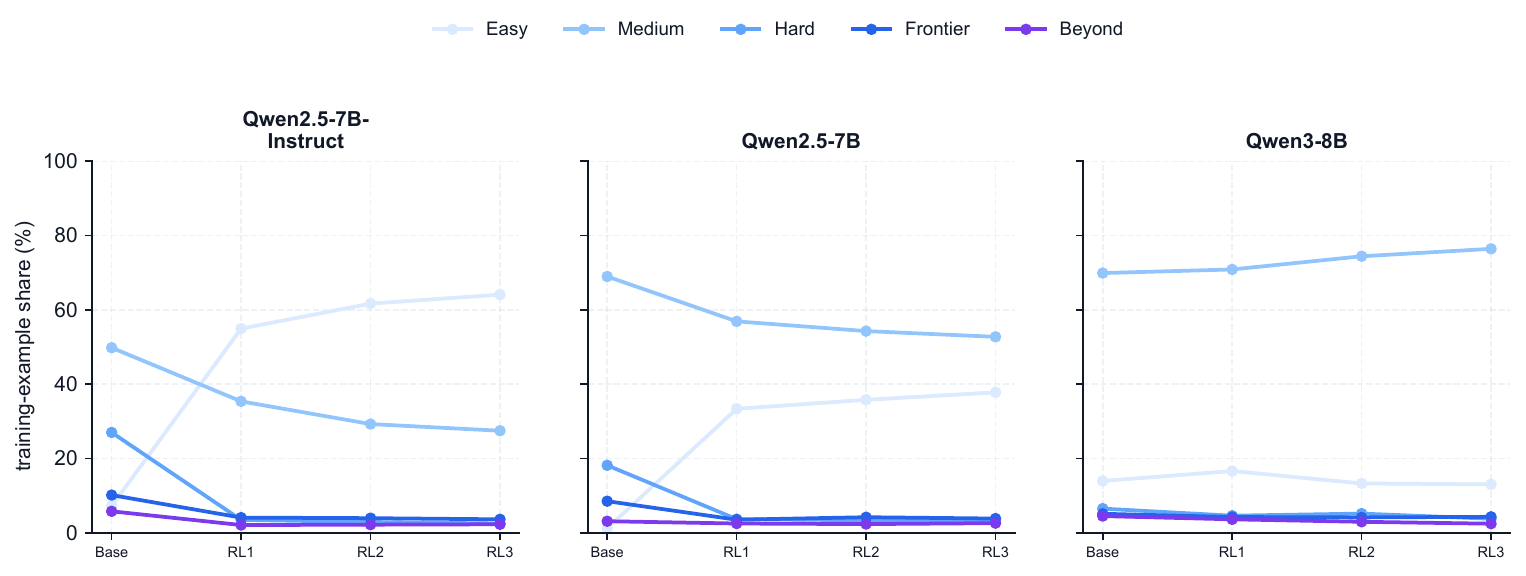}
  \caption{Training-example difficulty shares across curriculum rounds. We record the difficulty groups produced by pass@$k$ exploration during the curriculum and plot the normalized share of Easy, Medium, Hard, Frontier, and Beyond examples. Compared with the stacked counts in Figure~\ref{fig:eps-stacks}, this normalized view makes model settings with different starting distributions easier to compare.}
  \label{fig:training-difficulty-shares}
\end{figure}

\begin{figure}[p]
  \centering
  \includegraphics[width=\linewidth]{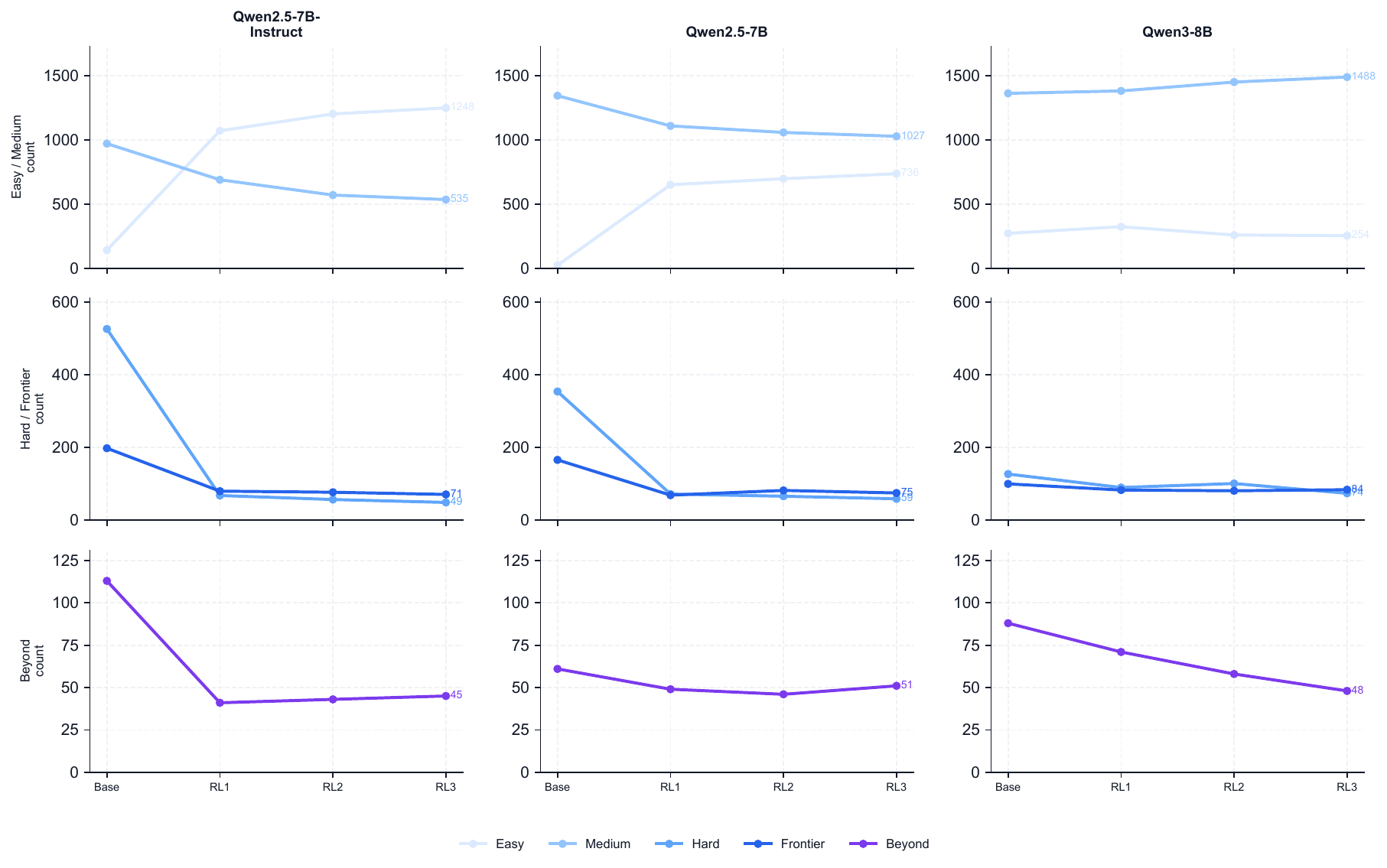}
  \caption{Absolute training-example difficulty counts across curriculum rounds. The top row separates Easy and Medium counts, the middle row separates Hard and Frontier counts, and the bottom row shows Beyond counts. This complements the normalized view by showing the scale of movement from Beyond into reward-reachable groups.}
  \label{fig:training-difficulty-counts}
\end{figure}

\begin{figure}[p]
  \centering
  \includegraphics[width=\linewidth]{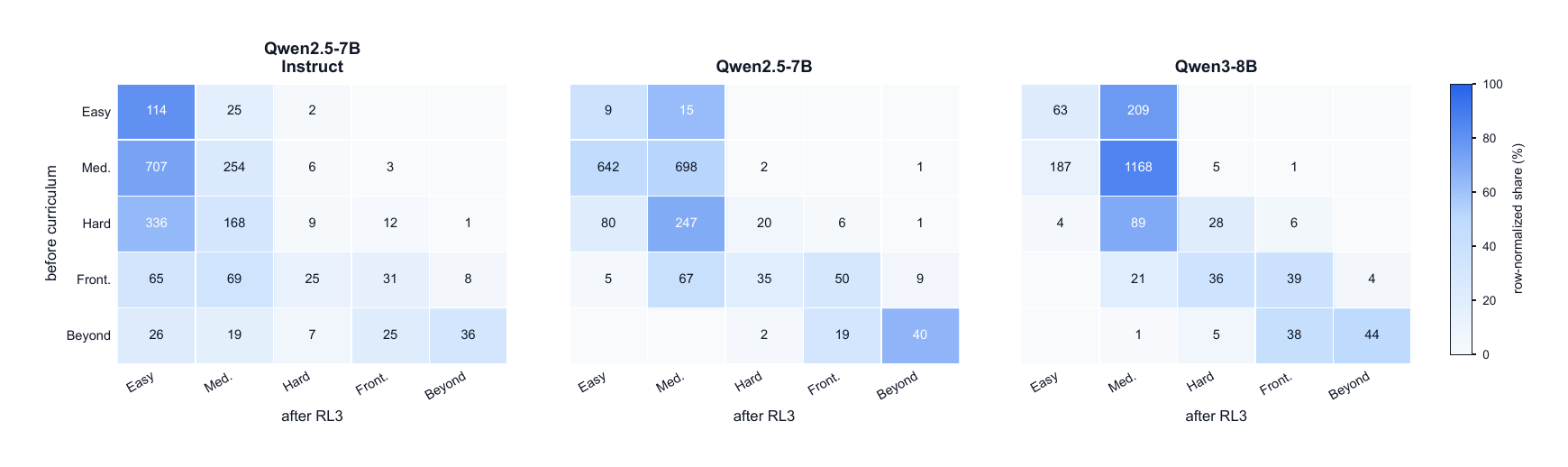}
  \caption{Training-example difficulty transitions from the base model to the third curriculum round. Rows denote the difficulty assigned before curriculum training, and columns denote the difficulty assigned after RL3. Shading is normalized within each starting difficulty group, and cell labels show counts. Movement from the Beyond row into Easy, Medium, Hard, or Frontier indicates examples that were not solved within the 256-sample exploration budget before curriculum training but became reward-reachable afterwards.}
  \label{fig:training-transition-matrices}
\end{figure}

This appendix provides additional visual evidence for the main empirical claims: boundary-aware Curriculum RL improves large-\(k\) behavior more consistently than Vanilla RLVR, and the curriculum expands the part of the training pool that can provide useful reward-driven signal. The figures are organized around two questions. Figures~\ref{fig:passk-curves} and~\ref{fig:pass256-gain-heatmap} examine whether the reported gains appear at large sampling budgets. Figures~\ref{fig:training-difficulty-shares}--\ref{fig:training-transition-matrices} examine how the curriculum changes the difficulty distribution of training examples.

Figure~\ref{fig:passk-curves} gives a closer view of representative model--benchmark pairs where large-\(k\) behavior is especially informative. The purpose is not to replace the full grid in Figure~\ref{fig:full-passk-grid}, but to make the curve shape easier to inspect. The conclusion is consistent with the main text: Vanilla RLVR can improve early sampling behavior, but the large-\(k\) endpoint is less stable; boundary-aware Curriculum RL more directly targets whether additional problems become solvable under the same sampling budget.

Figure~\ref{fig:pass256-gain-heatmap} summarizes the same question at the benchmark level. It shows that the average pass@256 improvement is not driven by a single benchmark: positive Curriculum RL gains appear across model families and evaluation sets. It also makes the contrast with Vanilla RLVR explicit, since Vanilla RLVR has both positive and negative pass@256 changes depending on the model--benchmark pair.

Figure~\ref{fig:training-difficulty-shares} focuses on the training examples used by the curriculum process. The normalized view shows whether the curriculum is merely improving a fixed set of easy examples or changing the distribution of difficulty groups. Across the recorded curriculum rounds, the Beyond share decreases and more examples fall into reward-reachable groups, supporting the mechanism that curriculum training expands the part of the training pool from which RL can obtain useful reward-driven signal.

Figure~\ref{fig:training-difficulty-counts} reports the same training-example dynamics in absolute counts. This is useful because normalized shares can hide the scale of the movement. The figure shows that the reduction in Beyond examples corresponds to a real increase in examples solved at least once within the 256-sample exploration budget, rather than only a relative change caused by redistribution among already reachable groups.

Figure~\ref{fig:training-transition-matrices} provides a per-example view of the same effect. The key evidence is movement from the Beyond row into Easy, Medium, Hard, or Frontier columns, which means that examples previously unsolved within the pass@$k$ exploration budget become reward-reachable after curriculum training. This transition-level view supports the interpretation that boundary-aware Curriculum RL changes the status of concrete examples, not only aggregate pass@$k$ averages.

Taken together, these supplementary figures support the same mechanism from two directions: the evaluation curves show that large-\(k\) behavior improves, and the training-example diagnostics show that more examples become available for reward-driven RL updates.
\FloatBarrier

\section{Implementation Details}
This section records implementation choices needed to reproduce the reported curriculum runs. The key design choice is the curriculum range used in each RL round; the teacher-guidance selection rule and prompt are given in Section~\ref{sec:appendix-sft-details}.

\subsection{Curriculum Choices}
\label{sec:appendix-curriculum-selection}

Table~\ref{tab:curriculum-choices} reports the curriculum range used in each RL round. The choices follow the Stage-III principle: use the hardest range that still contains enough reward-reachable examples for RL, then move the lower end of the range as the observed reasoning capacity boundary changes. The table should be read as an implementation record of the reported runs. Its main role is to make the curriculum process reproducible and to show how the simple boundary-aware rule is instantiated across model settings.

\begin{table}[!htbp]
\centering
\caption{Curriculum ranges used in the three reported curriculum rounds.}
\label{tab:curriculum-choices}
\vspace{2pt}
\footnotesize
\setlength{\tabcolsep}{4.6pt}
\renewcommand{\arraystretch}{1.12}
\begin{tabular}{@{}p{0.34\linewidth}p{0.18\linewidth}p{0.18\linewidth}p{0.18\linewidth}@{}}
\toprule
Model setting & RL1 & RL2 & RL3 \\
\midrule
Qwen2.5-7B-Instruct & Beyond-only & Frontier-to-Beyond & Easy-to-Beyond \\
Qwen2.5-7B & Medium-to-Beyond & Beyond-only & Frontier-to-Beyond \\
Qwen3-8B & Medium-to-Beyond & Medium-to-Beyond & Medium-to-Beyond \\
LLaMA-3.1-8B-Instruct & Medium-to-Beyond & Medium-to-Beyond & Beyond-only \\
DeepSeek-R1-0528-Qwen3-8B & Easy-to-Beyond & Frontier-to-Beyond & Frontier-to-Beyond \\
\bottomrule
\end{tabular}
\end{table}
\renewcommand{\arraystretch}{1}
\FloatBarrier

\section{Mechanistic Analysis of Zero-Advantage Groups}
\label{sec:appendix-zero-advantage}

This section formalizes the mechanism used in the main text. The analysis is intentionally local: it concerns the reward-driven term of a GRPO-style update with binary outcome rewards. It does not claim that all optimizer terms vanish, nor does it characterize every possible source of representation change. Its purpose is to make precise why all-failed rollout groups provide no task-specific preference signal, and why moving a boundary example into a reward-reachable regime changes the optimization problem faced by RL.

\paragraph{Setup.}
For a prompt \(x\), suppose the current policy samples a group of \(G\) rollouts \(\{y_i\}_{i=1}^G\) with binary rewards \(r_i\in\{0,1\}\). The group-normalized advantage is
\[
  \hat A_i = \frac{r_i-\bar r}{\sigma_r+\epsilon},
  \qquad
  \bar r = \frac{1}{G}\sum_{j=1}^G r_j ,
\]
where \(\sigma_r\) is the standard deviation of rewards within the group and \(\epsilon>0\). The reward-weighted policy-gradient term for this prompt has the form
\[
  g_x(\theta)=
  \sum_{i=1}^G \hat A_i \nabla_\theta \log \pi_\theta(y_i\mid x).
\]

\paragraph{Proposition 1: all-equal reward groups have zero reward weights.}
If all rollouts in the group receive the same reward, then \(\hat A_i=0\) for every rollout and \(g_x(\theta)=0\).

\paragraph{Proof.}
If all rewards are equal to a constant \(c\in\{0,1\}\), then \(\bar r=c\) and \(r_i-\bar r=0\) for every \(i\). Therefore \(\hat A_i=0\) for all sampled rollouts, and the weighted sum \(g_x(\theta)\) is zero. In particular, an all-failed group has \(c=0\) and contributes no positive or negative preference among its sampled trajectories. An all-success group also has zero within-group reward contrast; however, all-failed groups are the central boundary case because they arise when the current policy has not produced any successful trajectory from which outcome-reward RL can learn. \(\square\)

\paragraph{Proposition 2: mixed-reward groups are the only binary groups with nonzero reward contrast.}
Let \(m=\sum_{i=1}^G r_i\) be the number of successful rollouts. If \(0<m<G\), then successful and failed rollouts receive advantage weights with opposite signs. If \(m=0\) or \(m=G\), all advantage weights are zero.

\paragraph{Proof.}
For binary rewards, \(\bar r=m/G\) and
\[
  \sigma_r
  =
  \sqrt{\frac{1}{G}\sum_{i=1}^G (r_i-\bar r)^2}
  =
  \frac{\sqrt{m(G-m)}}{G}.
\]
When \(0<m<G\), a successful rollout has
\[
  \hat A^{+}
  =
  \frac{1-m/G}{\sqrt{m(G-m)}/G+\epsilon}
  >0,
\]
whereas a failed rollout has
\[
  \hat A^{-}
  =
  \frac{-m/G}{\sqrt{m(G-m)}/G+\epsilon}
  <0.
\]
Thus the group supplies an explicit reward contrast: successes are upweighted and failures are downweighted. When \(m=0\) or \(m=G\), Proposition~1 applies and all advantages are zero. Without the numerical stabilizer \(\epsilon\), the two nonzero weights reduce to \(\hat A^{+}=\sqrt{(G-m)/m}\) and \(\hat A^{-}=-\sqrt{m/(G-m)}\). \(\square\)

\paragraph{Proposition 3: probability of receiving reward contrast.}
Let \(p_\theta(x)\) be the probability that one rollout from the current policy solves \(x\) under a fixed sampler. Under the idealized assumption that the \(G\) rollouts in a group are independent, the probability that the group contains both at least one success and at least one failure is
\[
  q_G(p_\theta) = 1 - (1-p_\theta)^G - p_\theta^G .
\]
Moreover, \(q_G(p_\theta)=0\) at \(p_\theta=0\) and \(p_\theta=1\), \(q_G(p_\theta)>0\) for \(0<p_\theta<1\), and \(q_G\) is maximized at \(p_\theta=1/2\).

\paragraph{Proof.}
The group has no reward contrast in two cases: all rollouts fail, which occurs with probability \((1-p_\theta)^G\), or all rollouts succeed, which occurs with probability \(p_\theta^G\). The complement is the event that the group contains at least one success and at least one failure, giving the expression above. Its derivative is
\[
  q'_G(p_\theta)
  =
  G(1-p_\theta)^{G-1} - Gp_\theta^{G-1}.
\]
The derivative is positive when \(p_\theta<1/2\), zero at \(p_\theta=1/2\), and negative when \(p_\theta>1/2\). Hence \(q_G\) is maximized at \(p_\theta=1/2\). Near the zero-success regime, \(q_G(p_\theta)\approx Gp_\theta\) for small \(p_\theta\), so even a small increase in the probability of producing a correct trajectory can substantially increase the chance that an RL group contains usable reward contrast. \(\square\)

\paragraph{Corollary 1: zero observed successes mark a low-success-probability regime.}
The pass@$k$ exploration stage provides an empirical way to locate prompts with very small observed success probability under the current sampler. Under the same idealized independent-rollout model, if \(K\) samples produce no success, then a one-sided \(1-\alpha\) upper bound on \(p_\theta(x)\) is
\[
  p_\theta(x) \le 1-\alpha^{1/K}.
\]
For \(K=256\) and \(\alpha=0.05\), this gives
\[
  p_\theta(x) \le 1 - 0.05^{1/256} \approx 0.0116 .
\]

\paragraph{Proof.}
The probability of seeing no success in \(K\) independent samples is
\[
  \Pr[\text{no success in }K\text{ samples}]
  =
  (1-p_\theta(x))^K .
\]
Solving \((1-p_\theta(x))^K=\alpha\) for \(p_\theta(x)\) gives the stated upper bound. We use this calculation only as an interpretation of the pass@$k$ signal, not as a formal guarantee about dependent decoding samples. It clarifies why Beyond examples are likely to form zero-advantage groups during ordinary rollout training. \(\square\)

\paragraph{Corollary 2: making a boundary example reward-reachable is sufficient to create nonzero signal probability.}
If targeted guidance moves an example from \(p_\theta(x)=0\) to any \(p_\theta(x)\in(0,1)\), then the probability of drawing a mixed-reward group becomes strictly positive under the independent-rollout model.

\paragraph{Proof.}
By Proposition~3, \(q_G(0)=0\) and \(q_G(p_\theta)>0\) for every \(p_\theta\in(0,1)\). Therefore, once an example becomes reward-reachable but not trivially solved by every rollout, a group update can sample both successful and failed trajectories with positive probability. By Proposition~2, such a group has nonzero reward contrast. \(\square\)

\paragraph{Implication for boundary-aware curriculum RL.}
The mechanism can be summarized as
\begin{center}
\begin{tcolorbox}[
  enhanced,
  width=0.82\linewidth,
  colback=PaperBlue!3,
  colframe=PaperGray!70,
  boxrule=0.4pt,
  arc=1pt,
  left=4pt,
  right=4pt,
  top=2pt,
  bottom=2pt,
  boxsep=0pt,
  halign=center
]
\footnotesize
\(
  \text{zero-advantage group}
  \rightarrow
  \text{reward-reachable example}
  \rightarrow
  \text{RL reinforces new reasoning}.
\)
\end{tcolorbox}
\end{center}
Large-sample pass@$k$ exploration locates examples near the empirical reasoning capacity boundary. Targeted teacher guidance then moves a small number of selected examples away from the zero-success regime, making mixed-reward rollout groups possible. Once such reward contrast exists, RL can consolidate the newly introduced reasoning pattern rather than repeatedly sampling all-failed groups with zero reward-driven advantage.

\clearpage

\section{Supplementary Experimental Tables}
\label{sec:appendix-tables}
The tables below provide the numerical details behind the appendix figures and the main pass@$k$ results. They report only the base model, Vanilla RLVR model, and reported Curriculum RL model used in the paper, together with the training-example difficulty summaries recorded during the curriculum. Together, these tables support three conclusions: the teacher intervention is small and concentrated near the boundary; Curriculum RL improves pass@256 more consistently than Vanilla RLVR; and the training pool contains more reward-reachable examples after curriculum learning.

\subsection{Teacher-Guidance Seed Composition}
Table~\ref{tab:sft-composition} documents the scale of the teacher-guidance bridge. Each round uses only a small number of Frontier and Beyond examples, rather than a large teacher-generated corpus. This supports the design goal of using teacher guidance locally near the reasoning capacity boundary while leaving RL responsible for consolidating the introduced reasoning patterns.
\footnotesize
\setlength{\tabcolsep}{4.8pt}
\renewcommand{\arraystretch}{1.08}
\setlength{\LTleft}{0pt plus 1fill}
\setlength{\LTright}{0pt plus 1fill}
\begin{longtable}{p{0.38\linewidth}lrrr}
\caption{Teacher-guidance seed composition. Each round uses a small set of Frontier and Beyond examples selected near the current reasoning capacity boundary.}\label{tab:sft-composition}\\
\toprule
Model setting & Round & N & Frontier & Beyond \\
\midrule
\endfirsthead
\toprule
Model setting & Round & N & Frontier & Beyond \\
\midrule
\endhead
Qwen2.5-7B-Instruct & RL1 & 25 & 10 & 15 \\
Qwen2.5-7B-Instruct & RL2 & 35 & 15 & 20 \\
Qwen2.5-7B-Instruct & RL3 & 43 & 18 & 25 \\
Qwen2.5-7B & RL1 & 30 & 15 & 15 \\
Qwen2.5-7B & RL2 & 20 & 10 & 10 \\
Qwen2.5-7B & RL3 & 30 & 15 & 15 \\
Qwen3-8B & RL1 & 20 & 10 & 10 \\
Qwen3-8B & RL2 & 30 & 10 & 20 \\
Qwen3-8B & RL3 & 40 & 10 & 30 \\
LLaMA-3.1-8B-Instruct & RL1 & 20 & 10 & 10 \\
LLaMA-3.1-8B-Instruct & RL2 & 20 & 10 & 10 \\
LLaMA-3.1-8B-Instruct & RL3 & 20 & 10 & 10 \\
DeepSeek-R1-0528-Qwen3-8B & RL1 & 20 & 10 & 10 \\
DeepSeek-R1-0528-Qwen3-8B & RL2 & 20 & 10 & 10 \\
DeepSeek-R1-0528-Qwen3-8B & RL3 & 30 & 10 & 20 \\
\bottomrule
\end{longtable}
\normalsize

\subsection{Per-Benchmark pass@256 Gain Summary}
Table~\ref{tab:appendix-pass256-gains} expands the pass@256 comparison behind Figure~\ref{fig:pass256-gain-heatmap}. The table shows that Vanilla RLVR sometimes improves pass@256 but also produces several negative changes relative to the base model. Curriculum RL has broader positive gains, especially on AIME 2024 and AIME 2025, indicating that the large-\(k\) improvement is not restricted to the averaged metric in the main text.
\footnotesize
\setlength{\tabcolsep}{2.8pt}
\renewcommand{\arraystretch}{1.08}
\setlength{\LTleft}{0pt plus 1fill}
\setlength{\LTright}{0pt plus 1fill}
\begin{longtable}{p{0.31\linewidth}lrrr}
\caption{Per-benchmark pass@256 gains for Vanilla RLVR and Curriculum RL. Values are percentage-point differences computed relative to the indicated reference model.}\label{tab:appendix-pass256-gains}\\
\toprule
Model & Benchmark & \shortstack{Vanilla RLVR\\$-$ Base} & \shortstack{Curriculum RL\\$-$ Base} & \shortstack{Curriculum RL\\$-$ Vanilla RLVR} \\
\midrule
\endfirsthead
\toprule
Model & Benchmark & \shortstack{Vanilla RLVR\\$-$ Base} & \shortstack{Curriculum RL\\$-$ Base} & \shortstack{Curriculum RL\\$-$ Vanilla RLVR} \\
\midrule
\endhead
Qwen2.5-7B-Instruct & AIME 2024 & +3.3 & +13.3 & +10.0 \\
Qwen2.5-7B-Instruct & AIME 2025 & +0.0 & +10.0 & +10.0 \\
Qwen2.5-7B-Instruct & MATH500 & +0.2 & +0.8 & +0.6 \\
\midrule
Qwen2.5-7B & AIME 2024 & -13.3 & +12.2 & +25.6 \\
Qwen2.5-7B & AIME 2025 & +0.0 & +18.9 & +18.9 \\
Qwen2.5-7B & MATH500 & -1.7 & +1.7 & +3.5 \\
\midrule
Qwen3-8B & AIME 2024 & +13.3 & +26.7 & +13.3 \\
Qwen3-8B & AIME 2025 & +24.4 & +21.1 & -3.3 \\
Qwen3-8B & MATH500 & +4.7 & +3.1 & -1.6 \\
\midrule
LLaMA-3.1-8B-Instruct & AIME 2024 & -16.7 & +6.7 & +23.3 \\
LLaMA-3.1-8B-Instruct & AIME 2025 & -10.0 & +0.0 & +10.0 \\
LLaMA-3.1-8B-Instruct & MATH500 & -18.2 & +14.4 & +32.6 \\
\midrule
DeepSeek-R1-0528-Qwen3-8B & AIME 2024 & -3.3 & +3.3 & +6.7 \\
DeepSeek-R1-0528-Qwen3-8B & AIME 2025 & +3.3 & +6.7 & +3.3 \\
DeepSeek-R1-0528-Qwen3-8B & MATH500 & +6.0 & +7.4 & +1.4 \\
\bottomrule
\end{longtable}
\normalsize

\subsection{Supplementary pass@k Curve Values}
Table~\ref{tab:appendix-passk-curves} provides the numerical values for the full pass@$k$ curves. These values make it possible to inspect whether improvements appear only near pass@1 or persist as the sampling budget increases. The main pattern is that boundary-aware Curriculum RL often improves the curve at both small and large \(k\), whereas Vanilla RLVR is less stable at the pass@256 endpoint.
\fontsize{6.2}{7.1}\selectfont
\setlength{\tabcolsep}{1.45pt}
\renewcommand{\arraystretch}{1.07}
\setlength{\LTleft}{0pt plus 1fill}
\setlength{\LTright}{0pt plus 1fill}
\begin{longtable}{p{0.205\linewidth}p{0.095\linewidth}p{0.105\linewidth}rrrrrrrrr}
\caption{Pass@$k$ curve values for the base model, Vanilla RLVR model, and Curriculum RL curve shown in Figure~\ref{fig:full-passk-grid}. Values are percentages.}\label{tab:appendix-passk-curves}\\
\toprule
Model & Benchmark & Method & p@1 & p@2 & p@4 & p@8 & p@16 & p@32 & p@64 & p@128 & p@256 \\
\midrule
\endfirsthead
\toprule
Model & Benchmark & Method & p@1 & p@2 & p@4 & p@8 & p@16 & p@32 & p@64 & p@128 & p@256 \\
\midrule
\endhead
Qwen2.5-7B-Instruct & AIME 2024 & Base & 3.3 & 3.3 & 10.0 & 16.7 & 20.0 & 20.0 & 33.3 & 36.7 & 46.7 \\
Qwen2.5-7B-Instruct & AIME 2024 & Vanilla RLVR & 10.0 & 13.3 & 13.3 & 16.7 & 16.7 & 23.3 & 33.3 & 40.0 & 50.0 \\
Qwen2.5-7B-Instruct & AIME 2024 & Curriculum RL & 6.7 & 13.3 & 13.3 & 13.3 & 16.7 & 30.0 & 33.3 & 46.7 & 60.0 \\
Qwen2.5-7B-Instruct & AIME 2025 & Base & 0.0 & 0.0 & 3.3 & 16.7 & 23.3 & 30.0 & 30.0 & 36.7 & 53.3 \\
Qwen2.5-7B-Instruct & AIME 2025 & Vanilla RLVR & 3.3 & 3.3 & 6.7 & 13.3 & 16.7 & 26.7 & 33.3 & 40.0 & 53.3 \\
Qwen2.5-7B-Instruct & AIME 2025 & Curriculum RL & 6.7 & 10.0 & 10.0 & 16.7 & 20.0 & 30.0 & 43.3 & 53.3 & 63.3 \\
Qwen2.5-7B-Instruct & MATH500 & Base & 49.8 & 65.4 & 78.2 & 84.6 & 90.0 & 92.8 & 93.8 & 95.0 & 95.4 \\
Qwen2.5-7B-Instruct & MATH500 & Vanilla RLVR & 66.4 & 79.2 & 85.6 & 88.4 & 91.0 & 91.8 & 93.6 & 95.0 & 95.6 \\
Qwen2.5-7B-Instruct & MATH500 & Curriculum RL & 60.4 & 74.2 & 80.6 & 86.6 & 89.8 & 92.6 & 94.2 & 95.2 & 96.2 \\
\midrule
Qwen2.5-7B & AIME 2024 & Base & 1.1 & 2.2 & 5.6 & 5.6 & 11.1 & 13.3 & 21.1 & 30.0 & 40.0 \\
Qwen2.5-7B & AIME 2024 & Vanilla RLVR & 0.0 & 0.0 & 6.7 & 6.7 & 10.0 & 13.3 & 16.7 & 20.0 & 26.7 \\
Qwen2.5-7B & AIME 2024 & Curriculum RL & 6.7 & 8.9 & 12.2 & 16.7 & 21.1 & 25.6 & 31.1 & 43.3 & 52.2 \\
Qwen2.5-7B & AIME 2025 & Base & 0.0 & 1.1 & 1.1 & 2.2 & 4.4 & 8.9 & 10.0 & 14.4 & 26.7 \\
Qwen2.5-7B & AIME 2025 & Vanilla RLVR & 0.0 & 0.0 & 3.3 & 10.0 & 10.0 & 23.3 & 23.3 & 23.3 & 26.7 \\
Qwen2.5-7B & AIME 2025 & Curriculum RL & 2.2 & 5.6 & 8.9 & 11.1 & 18.9 & 23.3 & 31.1 & 40.0 & 45.6 \\
Qwen2.5-7B & MATH500 & Base & 17.7 & 30.3 & 46.8 & 63.1 & 76.1 & 83.9 & 88.1 & 91.5 & 94.1 \\
Qwen2.5-7B & MATH500 & Vanilla RLVR & 24.0 & 38.6 & 57.6 & 75.0 & 81.8 & 87.0 & 89.0 & 90.8 & 92.4 \\
Qwen2.5-7B & MATH500 & Curriculum RL & 46.1 & 63.9 & 74.8 & 82.5 & 87.5 & 90.9 & 92.4 & 94.3 & 95.9 \\
\midrule
Qwen3-8B & AIME 2024 & Base & 0.0 & 4.4 & 4.4 & 6.7 & 10.0 & 10.0 & 14.4 & 21.1 & 23.3 \\
Qwen3-8B & AIME 2024 & Vanilla RLVR & 3.3 & 3.3 & 6.7 & 13.3 & 20.0 & 26.7 & 30.0 & 30.0 & 36.7 \\
Qwen3-8B & AIME 2024 & Curriculum RL & 3.3 & 3.3 & 6.7 & 13.3 & 20.0 & 23.3 & 30.0 & 36.7 & 50.0 \\
Qwen3-8B & AIME 2025 & Base & 0.0 & 1.1 & 2.2 & 4.4 & 7.8 & 8.9 & 10.0 & 10.0 & 12.2 \\
Qwen3-8B & AIME 2025 & Vanilla RLVR & 10.0 & 10.0 & 10.0 & 10.0 & 20.0 & 20.0 & 23.3 & 30.0 & 36.7 \\
Qwen3-8B & AIME 2025 & Curriculum RL & 0.0 & 0.0 & 0.0 & 3.3 & 13.3 & 20.0 & 20.0 & 23.3 & 33.3 \\
Qwen3-8B & MATH500 & Base & 20.3 & 34.3 & 50.3 & 64.0 & 74.7 & 82.3 & 86.5 & 88.9 & 91.7 \\
Qwen3-8B & MATH500 & Vanilla RLVR & 46.2 & 66.0 & 80.6 & 88.0 & 90.6 & 92.2 & 94.4 & 95.6 & 96.4 \\
Qwen3-8B & MATH500 & Curriculum RL & 28.4 & 43.8 & 59.6 & 72.6 & 83.0 & 88.6 & 90.4 & 92.6 & 94.8 \\
\midrule
LLaMA-3.1-8B-Instruct & AIME 2024 & Base & 0.0 & 0.0 & 0.0 & 6.7 & 13.3 & 20.0 & 33.3 & 36.7 & 53.3 \\
LLaMA-3.1-8B-Instruct & AIME 2024 & Vanilla RLVR & 3.3 & 6.7 & 6.7 & 10.0 & 10.0 & 16.7 & 20.0 & 23.3 & 36.7 \\
LLaMA-3.1-8B-Instruct & AIME 2024 & Curriculum RL & 0.0 & 3.3 & 10.0 & 13.3 & 20.0 & 23.3 & 26.7 & 46.7 & 60.0 \\
LLaMA-3.1-8B-Instruct & AIME 2025 & Base & 0.0 & 0.0 & 0.0 & 3.3 & 6.7 & 6.7 & 6.7 & 13.3 & 20.0 \\
LLaMA-3.1-8B-Instruct & AIME 2025 & Vanilla RLVR & 0.0 & 0.0 & 0.0 & 0.0 & 0.0 & 0.0 & 3.3 & 10.0 & 10.0 \\
LLaMA-3.1-8B-Instruct & AIME 2025 & Curriculum RL & 0.0 & 3.3 & 3.3 & 6.7 & 13.3 & 13.3 & 13.3 & 16.7 & 20.0 \\
LLaMA-3.1-8B-Instruct & MATH500 & Base & 9.2 & 14.8 & 21.8 & 29.0 & 38.4 & 49.8 & 56.8 & 67.4 & 73.6 \\
LLaMA-3.1-8B-Instruct & MATH500 & Vanilla RLVR & 12.4 & 17.8 & 23.0 & 27.0 & 32.6 & 38.8 & 44.0 & 49.8 & 55.4 \\
LLaMA-3.1-8B-Instruct & MATH500 & Curriculum RL & 14.8 & 23.8 & 33.8 & 45.4 & 59.8 & 70.0 & 77.6 & 82.6 & 88.0 \\
\midrule
DeepSeek-R1-0528-Qwen3-8B & AIME 2024 & Base & 0.0 & 0.0 & 0.0 & 0.0 & 6.7 & 6.7 & 10.0 & 13.3 & 16.7 \\
DeepSeek-R1-0528-Qwen3-8B & AIME 2024 & Vanilla RLVR & 0.0 & 0.0 & 0.0 & 0.0 & 0.0 & 0.0 & 3.3 & 6.7 & 13.3 \\
DeepSeek-R1-0528-Qwen3-8B & AIME 2024 & Curriculum RL & 0.0 & 3.3 & 3.3 & 3.3 & 10.0 & 10.0 & 20.0 & 20.0 & 20.0 \\
DeepSeek-R1-0528-Qwen3-8B & AIME 2025 & Base & 3.3 & 3.3 & 3.3 & 3.3 & 3.3 & 6.7 & 10.0 & 10.0 & 10.0 \\
DeepSeek-R1-0528-Qwen3-8B & AIME 2025 & Vanilla RLVR & 0.0 & 0.0 & 3.3 & 3.3 & 3.3 & 3.3 & 6.7 & 10.0 & 13.3 \\
DeepSeek-R1-0528-Qwen3-8B & AIME 2025 & Curriculum RL & 0.0 & 0.0 & 3.3 & 3.3 & 6.7 & 6.7 & 10.0 & 13.3 & 16.7 \\
DeepSeek-R1-0528-Qwen3-8B & MATH500 & Base & 15.2 & 22.4 & 32.0 & 45.8 & 55.8 & 63.6 & 69.8 & 74.4 & 78.2 \\
DeepSeek-R1-0528-Qwen3-8B & MATH500 & Vanilla RLVR & 13.4 & 24.4 & 35.4 & 50.8 & 62.0 & 68.6 & 75.6 & 80.2 & 84.2 \\
DeepSeek-R1-0528-Qwen3-8B & MATH500 & Curriculum RL & 22.2 & 39.2 & 49.6 & 59.2 & 66.6 & 74.4 & 79.6 & 83.2 & 85.6 \\
\bottomrule
\end{longtable}
\normalsize

\subsection{Training-Example Signal Expansion Summary}
Table~\ref{tab:training-signal-expansion} summarizes the training-example difficulty changes that are visualized in Figure~\ref{fig:eps-stacks}. For the recorded Qwen settings, the number of Beyond examples decreases after curriculum training, while the number of examples solved at least once within the 256-sample budget increases. This supports the mechanism-level claim that the curriculum expands the region that can provide useful reward-driven signal to RL.
\footnotesize
\setlength{\tabcolsep}{3.6pt}
\begin{table}[!htbp]
\centering
\caption{Summary of the training-example difficulty changes shown in Figure~\ref{fig:eps-stacks}. Reachable denotes examples solved at least once within the 256-sample pass@$k$ exploration budget.}
\label{tab:training-signal-expansion}
\vspace{2pt}
\renewcommand{\arraystretch}{1.12}
\begin{tabular}{p{0.28\linewidth}rrrrrr}
\toprule
Model setting & \shortstack{Base\\Beyond} & \shortstack{RL3\\Beyond} & \shortstack{\(\Delta\)\\Beyond} & \shortstack{Base\\reachable} & \shortstack{RL3\\reachable} & \shortstack{\(\Delta\)\\reachable} \\
\midrule
Qwen2.5-7B & 61 & 51 & -10 & 1887 & 1897 & +10 \\
Qwen2.5-7B-Instruct & 113 & 45 & -68 & 1835 & 1903 & +68 \\
Qwen3-8B & 88 & 48 & -40 & 1860 & 1900 & +40 \\
\bottomrule
\end{tabular}
\end{table}
\normalsize
\renewcommand{\arraystretch}{1}

\subsection{Training-Example Difficulty Transition Summary}
Table~\ref{tab:training-transition-summary} gives a more detailed transition summary for Figure~\ref{fig:training-transition-matrices}. The most important column is Base-Beyond-to-reachable: it counts examples that were outside the observed reasoning capacity boundary before curriculum training but became reward-reachable by RL3. The positive counts in this column show that the observed boundary shift is caused by concrete example-level transitions.
\scriptsize
\setlength{\tabcolsep}{1.9pt}
\begin{table}[!htbp]
\centering
\caption{Summary of the per-example difficulty transitions shown in Figure~\ref{fig:training-transition-matrices}. Base-Beyond-to-reachable counts training examples that were Beyond before curriculum training but became Easy, Medium, Hard, or Frontier after RL3. New Beyond counts examples that were not Beyond before curriculum training but became Beyond after RL3. The capped difficulty index \(\tilde{k}\) equals the mean first-success index for solved examples and 257 for Beyond examples.}
\label{tab:training-transition-summary}
\vspace{2pt}
\renewcommand{\arraystretch}{1.05}
\begin{tabular}{@{}p{0.245\linewidth}rrrrrr@{}}
\toprule
Model setting & \shortstack{Base\\Beyond} & \shortstack{Base-Beyond\\reachable} & \shortstack{Base-Beyond\\remaining} & \shortstack{New\\Beyond} & \shortstack{Net\\Beyond} & \shortstack{Mean \(\tilde{k}\)\\Base \(\rightarrow\) RL3} \\
\midrule
Qwen2.5-7B-Instruct & 113 & 77 & 36 & 9 & -68 & 29.93 \(\rightarrow\) 10.92 \\
Qwen2.5-7B & 61 & 21 & 40 & 11 & -10 & 20.08 \(\rightarrow\) 12.49 \\
Qwen3-8B & 88 & 44 & 44 & 4 & -40 & 19.31 \(\rightarrow\) 13.51 \\
\bottomrule
\end{tabular}
\end{table}
\normalsize
\renewcommand{\arraystretch}{1}

\subsection{Training-Example Difficulty Summaries}
Table~\ref{tab:training-difficulty-complete} reports the full difficulty counts for each recorded curriculum round. These values complement the transition summary by showing the round-by-round trajectory of Easy, Medium, Hard, Frontier, and Beyond examples. The consistent reduction in mean capped difficulty and the increase in reachable examples provide additional evidence that the curriculum changes the training distribution toward more usable RL signal.
\footnotesize
\setlength{\tabcolsep}{3.2pt}
\renewcommand{\arraystretch}{1.08}
\setlength{\LTleft}{0pt plus 1fill}
\setlength{\LTright}{0pt plus 1fill}
\begin{longtable}{p{0.28\linewidth}lrrrrrrr}
\caption{Training-example difficulty summaries recorded during the reported curriculum rounds. Reachable denotes examples solved at least once within the 256-sample pass@$k$ exploration budget.}\label{tab:training-difficulty-complete}\\
\toprule
Model setting & Round & Easy & Medium & Hard & Frontier & Beyond & Reachable & Mean k \\
\midrule
\endfirsthead
\toprule
Model setting & Round & Easy & Medium & Hard & Frontier & Beyond & Reachable & Mean k \\
\midrule
\endhead
Qwen2.5-7B & Base & 24 & 1343 & 354 & 166 & 61 & 1887 & 12.43 \\
Qwen2.5-7B & RL1 & 650 & 1108 & 72 & 69 & 49 & 1899 & 5.64 \\
Qwen2.5-7B & RL2 & 697 & 1057 & 66 & 82 & 46 & 1902 & 6.10 \\
Qwen2.5-7B & RL3 & 736 & 1027 & 59 & 75 & 51 & 1897 & 5.91 \\
Qwen2.5-7B-Instruct & Base & 141 & 970 & 526 & 198 & 113 & 1835 & 15.94 \\
Qwen2.5-7B-Instruct & RL1 & 1070 & 689 & 68 & 80 & 41 & 1907 & 5.84 \\
Qwen2.5-7B-Instruct & RL2 & 1201 & 570 & 57 & 77 & 43 & 1905 & 5.40 \\
Qwen2.5-7B-Instruct & RL3 & 1248 & 535 & 49 & 71 & 45 & 1903 & 5.10 \\
Qwen3-8B & Base & 272 & 1361 & 127 & 100 & 88 & 1860 & 8.06 \\
Qwen3-8B & RL1 & 324 & 1380 & 90 & 83 & 71 & 1877 & 6.64 \\
Qwen3-8B & RL2 & 259 & 1449 & 101 & 81 & 58 & 1890 & 7.32 \\
Qwen3-8B & RL3 & 254 & 1488 & 74 & 84 & 48 & 1900 & 7.35 \\
\bottomrule
\end{longtable}
\normalsize
\renewcommand{\arraystretch}{1}

Overall, the supplementary tables provide the numerical basis for the appendix figures. They show that the teacher-guidance stage remains small, that pass@256 gains are broader under Curriculum RL than under Vanilla RLVR, and that the recorded training examples shift toward reward-reachable difficulty groups over curriculum rounds.

\section{Teacher-Guidance Details}
\label{sec:appendix-sft-details}

This section gives the sample-selection procedure and prompt template used to generate the small set of structured teacher traces.

\subsection{Boundary-Aware Sample Selection}
\label{sec:appendix-sft-selection}

Algorithm~\ref{alg:sft-selection} summarizes the sample-selection procedure. The input is the current pass@$k$ exploration result for the training examples. The procedure focuses on examples near the observed reasoning capacity boundary: Frontier examples are hard but still reachable within 256 samples, while Beyond examples remain unsolved within the 256-sample exploration budget.

\begin{algorithm}[!htbp]
\caption{Boundary-aware teacher-guidance sample selection}
\label{alg:sft-selection}
\small
\begin{algorithmic}[1]
\Require pass@$k$ exploration table \(D\), target counts \(n_F\) and \(n_B\)
\Ensure teacher-guidance seed set \(S\)
\State Keep examples with complete question, answer, problem-family tag, and pass@$k$ statistics.
\State Build Frontier candidate set \(C_F\) from examples labeled Frontier.
\State Build Beyond candidate set \(C_B\) from examples labeled Beyond.
\State Rank \(C_F\) by closeness to the Frontier threshold.
\State Rank \(C_B\) by closeness to the Beyond threshold.
\State Select \(n_F\) Frontier and \(n_B\) Beyond examples while keeping coverage across problem families balanced.
\State Remove examples already used in previous teacher-guidance rounds.
\State \Return the selected examples with their difficulty group and pass@$k$ statistics.
\end{algorithmic}
\end{algorithm}

\subsection{Prompt for Structured Teacher Traces}
\label{sec:appendix-sft-prompt}

The prompt asks the teacher model to provide a structured solution rather than only a final answer. It encourages a clear reasoning chain, a midpoint verification step, an alternative perspective, and a final consistency check.

\begin{tcolorbox}[
    enhanced,
    width=0.96\linewidth,
    colback=PaperBlue!3,
    colframe=black,
    boxrule=0.8pt,
    arc=1mm,
    left=6mm,
    right=6mm,
    top=4mm,
    bottom=4mm,
    fontupper=\small,
    title={Prompt for structured teacher traces},
    coltitle=white,
    colbacktitle=black,
    boxed title style={colback=black,colframe=black,arc=1mm,boxrule=0pt},
    attach boxed title to top left={xshift=4mm,yshift=-2mm}
]
Assume that you are a helpful mathematical reasoning assistant. You will receive a math problem. Your task is to provide a structured solution trace that can be used as supervised fine-tuning data for improving long-chain reasoning.

\vspace{0.4em}
\textbf{Logical backbone.}
Establish a clear deductive chain where each step justifies the next. Avoid unexplained jumps.

\vspace{0.4em}
\textbf{Explicit self-reflection.}
At approximately the midpoint of the reasoning, pause to verify the setup, arithmetic, transformation, or theorem choice before continuing.

\vspace{0.4em}
\textbf{Alternative-path exploration.}
Briefly explore at least one alternative perspective, method, or edge case before confirming the primary path.

\vspace{0.4em}
\textbf{Logical coherence check.}
Before providing the final answer, briefly summarize why the reasoning chain is consistent and why the final result follows.

\vspace{0.4em}
\textbf{Output requirement.}
Put the final answer in \verb|\boxed{}|. Do not omit the final answer.

\vspace{0.4em}
\textbf{Problem:} \{problem\}
\end{tcolorbox}
\normalsize


\end{document}